\documentclass{article}
\usepackage{ijcai24}

\usepackage{times}
\usepackage{soul}
\usepackage{url}
\usepackage[hidelinks]{hyperref}
\usepackage[utf8]{inputenc}
\usepackage[small]{caption}
\usepackage{graphicx}
\usepackage{amsmath}
\usepackage{amsthm}
\usepackage{booktabs}
\usepackage{algorithm}
\usepackage{algorithmic}
\usepackage[switch]{lineno}

\usepackage{amssymb}
\usepackage{multirow}
\usepackage{bm}
\usepackage{colortbl}
\usepackage{xcolor}
 \definecolor{verylightgray}{gray}{0.9}

\newtheorem{theorem}{Theorem}

\newtheorem{definition}{Definition}
\newtheorem{remark}{Remark}

\title{Weak Robust Compatibility Between Learning Algorithms and Counterfactual Explanation Generation Algorithms}

\author{
    Ao Xu$^1$
    \and
    Tieru Wu$^2$
    \affiliations
    Jilin University$^1$
    Jilin University$^2$
    \emails
    xuao22@mails.jlu.edu.cn,
    wutr@jlu.edu.cn
}

\begin{document}

\maketitle

\begin{abstract}
     Counterfactual explanation generation is a powerful method for  Explainable Artificial Intelligence. It can help users understand why machine learning models make specific decisions, and how to change those decisions. Evaluating the robustness of counterfactual explanation algorithms is therefore crucial.
    Previous literature has widely studied the robustness based on the perturbation of input instances. However, the robustness defined from the perspective of perturbed instances is sometimes biased, because this definition ignores the impact of learning algorithms on robustness.
    In this paper, we propose a more reasonable definition, Weak Robust Compatibility,  based on the perspective of explanation strength.
    In practice, we propose WRC-Test to help us generate more robust counterfactuals. Meanwhile, we designed experiments to verify the effectiveness of WRC-Test.
    Theoretically, we introduce the concepts of PAC learning theory and define the concept of PAC WRC-Approximability. Based on reasonable assumptions, we establish oracle inequalities about weak robustness, which gives a sufficient condition for PAC WRC-Approximability.
\end{abstract}

\section{Introduction}
\label{Intro}
In recent decades, the field of Artificial Intelligence (AI) has undergone significant transformations \cite{winston1984artificial,russell2010artificial}. The substantial expansion of available computing resources has spurred a new wave of research into AI algorithms, capturing considerable attention from industry stakeholders \cite{AttaranDeb2018}. Diverse algorithms and models have been developed in the AI domain, yielding decision-making outcomes of unprecedented precision. However, comprehending the decision-making mechanisms of many AI algorithms, particularly the black box algorithms endowed with millions of parameters, often proves daunting for users, rendering it nearly insurmountable to intuitively grasp their fundamental essence and characteristics. In essence, the outputs of these algorithms lack reliable explanations. This deficiency can erode trust in AI algorithms, thereby diminishing their practical utility. 
As a result, there has been a growing emphasis on research in interpretable artificial intelligence, as evidenced by the significant attention it has garnered in recent studies 
\cite{ghosh2020interpretable,doshi2017towards}.

\textbf{Explainable Artificial Intelligence (XAI).}
XAI is an interdisciplinary field that seeks to understand the decision-making processes and underlying principles of AI systems, and to convey this understanding to users. 
Markus et al. \cite{markus2021role} proposed a definition of Explainability and its properties. The paper also provides a comprehensive literature review to guide the design and evaluation of explainable AI systems. 
By identifying the strengths and weaknesses of AI systems in making decisions, it becomes possible to enhance the performance of existing models \cite{machlev2022}. 
XAI also plays a crucial role in building transparent and trustworthy systems, particularly in critical applications such as medical decision-making. 
Despite the development of numerous explicability techniques aimed at integrating interpretability into artificial intelligence solutions, the credibility of these explanations remains an enigmatic facet. 
Hence, researchers are currently engaged in extensive exploration of the diverse spectrum of XAI techniques. This pursuit underscores the significance and the dynamic nature of this research domain, which remains both crucial and continuously evolving \cite{xu2019explainable}. 

\textbf{Counterfactual Explanation.}
 Within the domain of XAI, Counterfactual Explanation Generation, as showcased in the study by Wachter et al. \cite{wachter2017counterfactual}, emerges as a powerful and effective method. It utilizes examples to provide users with a clear understanding of how artificial intelligence models make decisions \cite{stepin2021survey}. Its fundamental concept is rooted in counterfactual reasoning, where variations in input are observed to determine corresponding changes in output, thereby explaining real-world decisions. This approach, by presenting alternative scenarios, elucidates the decision-making process of artificial intelligence models, enhancing the transparency and credibility of their decisions. This, in turn, aids users in better comprehending and monitoring the operation of artificial intelligence systems.
For instance, consider an artificial intelligence model employed to determine whether a loan application should be approved or rejected. Empirical evidence suggests that \cite{guidotti2022counterfactual}, with an appropriate Counterfactual Explanation Generation Algorithm (CEGA), the generated counterfactuals can reveal how the decision would have been influenced if the applicant had a higher income or a longer credit history. By providing these alternative scenarios, counterfactual explanations empower users to gain deeper insights into the decision-making process of artificial intelligence models and identify potential latent biases or errors.
However, within the post-hoc paradigm \cite{Laugel2019}, when prior knowledge about ground-truth data and classifiers is lacking, the counterfactual examples generated by the classifier are susceptible to the influence of classifier robustness. For instance, if the classifier is already overfitting, the resulting counterfactual explanations may have limited value. Therefore, an important question arises: Given a CEGA and a Learning Algorithm (LA), 
how can one systematically assess the robustness of the generated counterfactual explanations in a more reasoned manner?

\textbf{Our Paper's Contributions.} In this work, we make the following contributions:
\begin{itemize}
    \item Conceptually, we propose a new notion of weak robust compatibility between CEGAs and LAs in Section \ref{Fc}. This notion takes into account the characteristics of both CEGAs and LAs and is more reasonable than the previous views in the literature.
    \item In terms of applications, we propose WRC-Test in Section \ref{Fc} to help generate more robust counterfactual explanations. We also conduct experiments in Section \ref{sec:exp} to validate the effectiveness of WRC-Test.
    \item From a theoretical perspective, we introduce the concept of PAC WRC-Approximability in Section \ref{M}. Based on reasonable assumptions, we establish oracle inequalities for weak robustness, which provides a sufficient condition for PAC WRC-Approximability. This result gives us a deeper understanding of robustness in the counterfactual explanation field.
\end{itemize}
\section{Related works}
\label{Rw}
Alvarez Melis et al. \cite{alvarez2018towards} introduced a quantitative robustness metric for explanation generation models in the context of neural network models. Their approach is straightforward: if a slight perturbation to an instance results in only minor disturbances to the corresponding counterfactual explanations, then the counterfactual explanations are deemed relatively robust. Subsequently, Alexander Levine et al. \cite{levine2019} extended this elementary idea by introducing stochastic smoothing techniques to enhance the robustness of explanations for neural network models. They provided theoretical justification for the introduced approach. Ahmad Ajalloeian et al. \cite{ajalloeian2022smoothed} conducted a comprehensive and systematic assessment of the robustness that smoothing techniques confer to explanations. Meanwhile, Marco Virgolin et al. 
 \cite{Virgolin2023} proposed two feasible definitions of robustness, each corresponding to different perturbation scenarios: one definition constrains the features that can be altered, while the other definition imposes constraints on the features that must remain unchanged. Their experiments demonstrate the crucial role of robustness of CEGAs for mitigating adverse perturbations. Furthermore, their experimental results indicate that the robustness defined by the first perturbation scenario is more reliable and computationally efficient compared to the robustness defined by the second perturbation scenario. In summary, these studies have contributed valuable insights into the robustness of CEGAs.

\section{Preliminaries}
\label{Pre}
\subsection{Learning Theory}\label{Pre_LT}
Let $\mathcal{X}$ be an instance space, $\mathcal{D}$ is a distribution over $\mathcal{Z}=\mathcal{X}\times \{-1,1\}$. Meanwhile, we write $\mathcal{H}$ as the hypothesis space, a set of classifiers from $\mathcal{X}$ to $\{-1,1\}$, $z=(x,y)\in \mathcal{Z}$ as a labeled sample, $\mathcal{Z}^m=\{z_1,z_2,..,z_m\}$ as a training sample with size $m$. 
Marginal distribution $\mathcal{D}_{\mathcal{X}}$ has a density $p(\textbf{x})$. We assume that $p(\textbf{x})\in [a,b]$, where $a > 0$, although the results of this study can be readily extended to the case where $a = 0$.
The error of a hypothesis $h$ according to $\mathcal{D}$ is 
\begin{align*}
R_{\mathcal{D}}(h)=\text{Pr}_{(x,y)\sim \mathcal D} (f(x)\ne y)=\mathbb E[\mathbb I(f(x)\ne y)].
\end{align*}
In learning theory, a learning algorithm $L$ for $\mathcal{H}$ is a function \cite{dreyfus2005neural}
$$L:\bigcup_{m=1}^{\infty} \mathcal{Z}^{m} \rightarrow \mathcal{H}$$
from the set of all training samples to $\mathcal{H}$, with properties: given $\forall \varepsilon\in(0,1),\forall \delta\in(0,1)$ and arbitrary distribution $\mathcal{D}$ over $\mathcal{Z}$,there is an integer $m(\varepsilon,\delta)$ such that if $m\geq m(\varepsilon,\delta)$,
$$R_{\mathcal D}(L(\mathcal{Z}^m))\leq R_{\mathcal{D}}(h^*)+\varepsilon$$ 
holds with probability at least $1-\delta$, where $h^*:=\text{argmin}_{h\in\mathcal H}R_\mathcal D(h)$. We typically refer to $h^*$ as a Bayesian optimal classifier w.r.t the current classification task. 
This type of result will be refered to as oracle inequality \cite{lerasle2019}. We define $R_{\mathcal{D}}(h)-R_{\mathcal{D}}\left(h^{*}\right)$  as the excess risk of $h$. In machine learning, excess risk is a metric used to measure the deviation between the performance of a classifier and that of the optimal classifier. 

\subsection{Counterfactual Explanations} \label{sec:PreCE}
A counterfactual explanation for $x\in\mathcal X,h\in \mathcal H$ is the minimal perturbation $\Delta _x$ of $x$ in the case of changing the classification result of $f$. Assuming that $d(\cdot,\cdot)$ is a differentiable distance function on $\mathcal X$, we define the optimal counterfactual example as
\begin{align}
\bar x^*:= \underset{\bar x }{\operatorname{argmin }} \    d(x,\bar x)\text{ s.t. }h(x)\neq h(\bar x).\label{eqdefCE}
\end{align}
The corresponding optimal counterfactual explanation is $$\Delta_x^*=\bar x^*-x.$$

The aforementioned definition is intuitively straightforward: the constraint of minimizing the distance between $x$ and $\bar{x}$ encourages $\bar{x}$ to closely align with the classification boundary of model $f$. This implies that the optimal counterfactual explanation, in some sense, reveals how to alter predictions with minimal effort, thus providing the user with easily understandable insights that may guide actions in comprehending the predictions made by model $f$. Meanwhile, we can consider the quantity $d(x,\bar x^* )$, which also carries its significance: it can be understood as the  counterfactua explanatory strength of a counterfactual explanation of $\bar x^*$ w.r.t $x$. It is noteworthy that interpreting $d(x, \bar x^*)$ as the  counterfactual \textbf{explanatory strength} of counterfactual explanations is a highly significant concept. We will leverage this simple yet profound concept to introduce the central concept of this paper, as discussed in Section \ref{subsecWRC}, referred to as Definition \ref{defWRC}.

\section{Weak Robust Compatibility and WRC-Test}
\label{Fc}
\subsection{The Mathematical Formulation of CEGAs}
For the sake of notational convenience, we adopt a more mathematical notation to define the CEGA. Similar to the definition of LAs, we define a CEGA as a mapping:
\begin{definition}[CEGA]\label{defCEGA}
Given a hypothesis space $\mathcal{H}$ and a instance space $\mathcal{X}$, we define a CEGA  $\mathcal{C}$ to be a mapping:
$$ \mathcal{C}:\mathcal{H}\times \mathcal{X}\rightarrow \mathcal{C},(h,x) \mapsto \mathcal{C}(h,x)$$
which maps $x$ to a counterfactual explanation of $x$ w.r.t the classifier $h$.
\end{definition}
In general, the mapping in Definition \ref{defCEGA} is typically determined by a metric, denoted as $d(\cdot,\cdot)$, following the form specified in Equation (\ref{eqdefCE}). We define such a CEGA as induced by the metric $d(\cdot,\cdot)$:
\begin{definition}[CEGAs induced by $d(\cdot,\cdot)$]\label{defCEGAd}
Given a hypothesis space $\mathcal{H}$, a instance space $\mathcal{X}$ and a metric $d(\cdot,\cdot)$ on $\mathcal{X}$. We refer to the following CEGA $\mathcal{C}_d$ as induced by $d(\cdot,\cdot)$:
\begin{align*}
\mathcal{C}_d: \ \mathcal{H} \times \mathcal{X}  &\rightarrow  \mathcal{X}\\
  (h, x) & \mapsto   \underset{ x' \in \mathcal{N} (x) }{\operatorname{argmin }} \ d (x',x)
\end{align*}
where $z = (x, y)$,
$\mathcal{N} (x) = \{ y \in \mathcal{X}: h (x) \neq h (y) \}$.
\end{definition}

In the following exposition, we will denote the instance space $\mathcal{X}$ as the metric space $(\mathcal{X}, d)$ for brevity. The choice of metric $d(\cdot,\cdot)$ offers numerous options, such as the Euclidean distance, Mahalanobis distance \cite{kanamori2020dace}, etc. No matter what metric $d(\cdot,\cdot)$ is chosen, there generally exist values for $C_U\geq C_L>0$ such that the following equation holds true: 
\begin{align}
    \forall x,y\in\mathcal{X}, C_L\cdot d(x,y)\leq d_E(x,y)\leq C_U\cdot d(x,y).
    \label{eqd}
\end{align}
where $d_E(\cdot,\cdot)$ represents the Euclidean distance. In this work, we consistently consider CEGAs as described in Definition \ref{defCEGAd}, and assume Equation (\ref{eqd}) is valid.
\begin{remark}
    It is noteworthy that we define CEGAs as mappings rather than functions. This is because, for a given instance $x$, the best counterfactual examples generated by CEGAs may not be unique. However, this multi-valued nature does not affect the reasoning in the subsequent context.
\end{remark}

\subsection{Defining the Robust Compatibility Between LAs and CEGAs}\label{subsecWRC}
\begin{figure}
    \centering
    \includegraphics[width=1.05\linewidth]{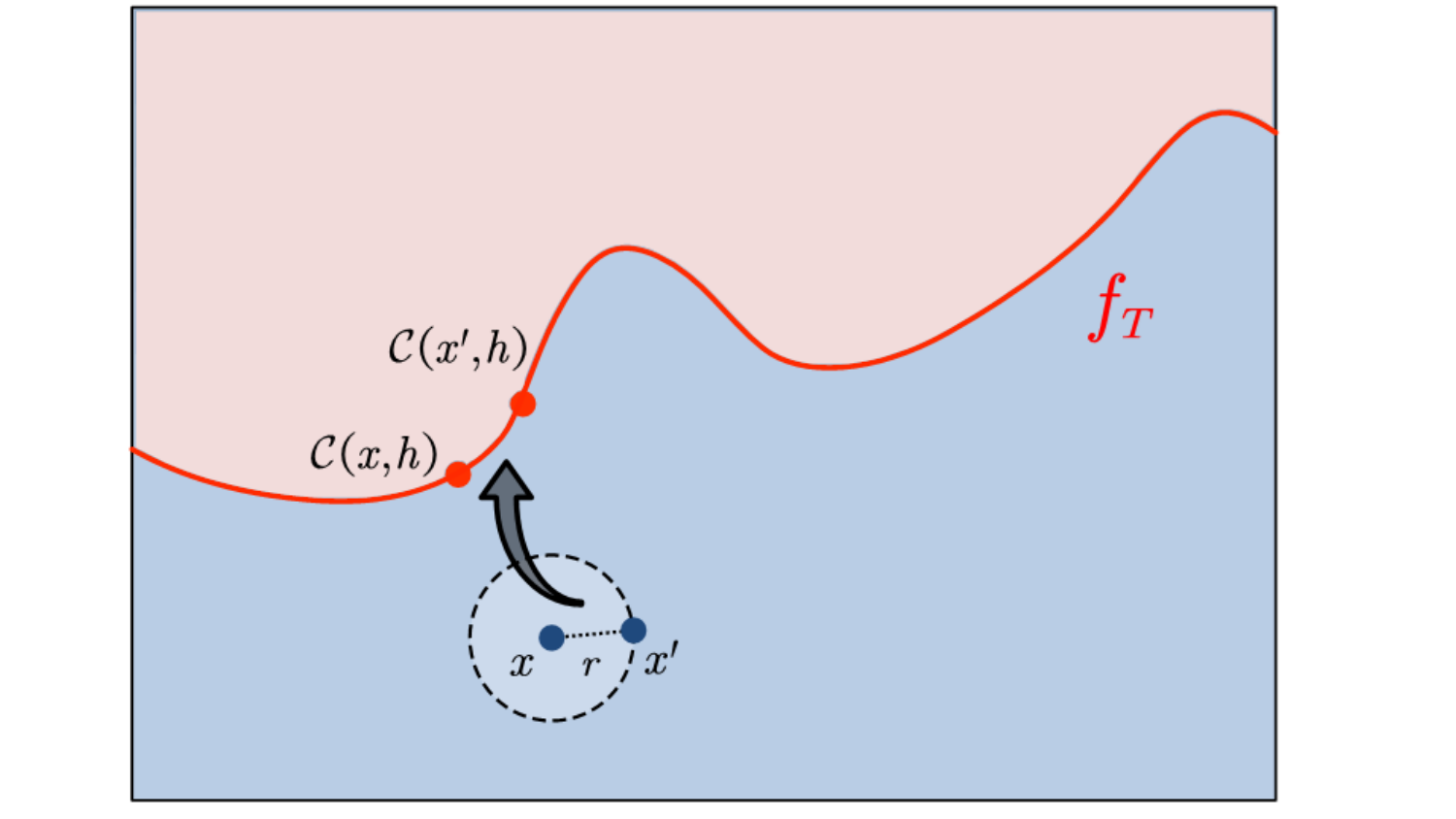}
    \caption{Illustration of the intuition behind the notion of SRC, where $f_T$ represents the classification boundary of classifier $h_T$.}
    \label{fig:SRC}
\end{figure}
In this section, we will present two definitions pertaining to robustness.  We denote the family of functions: $$\Phi = \{ \varphi : [0,+\infty) \rightarrow \mathbb{R}: \varphi (x) > 0,
\varphi'(x) < 0 ,\forall x\in
\mathbb R_+\}$$
and introduce a parameter $r>0$. To begin with, we furnish the inaugural definition.

\begin{definition}[Strong Robust Compatibility] \label{defSRC}
    Consider an instance space $(\mathcal{X},d)$, where  $\mathcal{X}$ is a subset of  $K$-dimensional Euclidean space $\mathbb{R}^K$, a LA represented as $L$, a CEGA as $\mathcal{C}$, $\varphi\in \Phi$.
    Let $D_T$ be a dataset of size $T$ drawn from the distribution $\mathcal{D}^{\otimes T}$, and $L(D_T)=h_T$.  We define the Strong Robust Compatibility (SRC) of  $\mathcal{C}$ w.r.t $h_T$ at $x$ as follows:
    \begin{align*}
    \text{SRC}_x^{\varphi} (\mathcal{C}, h_T) = \int_{
    \mathcal{X} \cap B (x, r)}\Delta(x,y)\cdot \varphi (d (x, y)) d y ,
    \end{align*}
     where $B(x, r)$ denotes a hypersphere centered at $x $ with a radius of $r$ and
     \begin{align*}
        \Delta(x,y)=d (\mathcal{C} (h_T, x), \mathcal{C} (h_T,
        y)).
     \end{align*} 
\end{definition}
\begin{remark}
    In our proposed definition, although we refer to SRC in terms of the definitions of CEGA $\mathcal{C}$ and classifier $h$, it is important to note that the classifier itself is determined by a learning algorithm $L$. The choice of the learning algorithm $L$ plays a pivotal role in determining essential properties of the boundaries and smoothness of the classifier obtained through training. Thus, we categorize SRC as a concept related to LAs and CEGAs.
\end{remark}
\begin{remark}
    Definition \ref{defSRC} abstracts the discussions of robustness in the past literature. For example, if we set $\varphi=1/x$ and $d(x,y)=\|x-y\|_2$ in Definition \ref{defSRC}, then we obtain the familiar notion of robustness 
 (The absence of  $\varphi$ from $\Phi$ does not affect the essential nature).
  
\end{remark}
\begin{figure}
    \centering
    \includegraphics[width=1\linewidth]{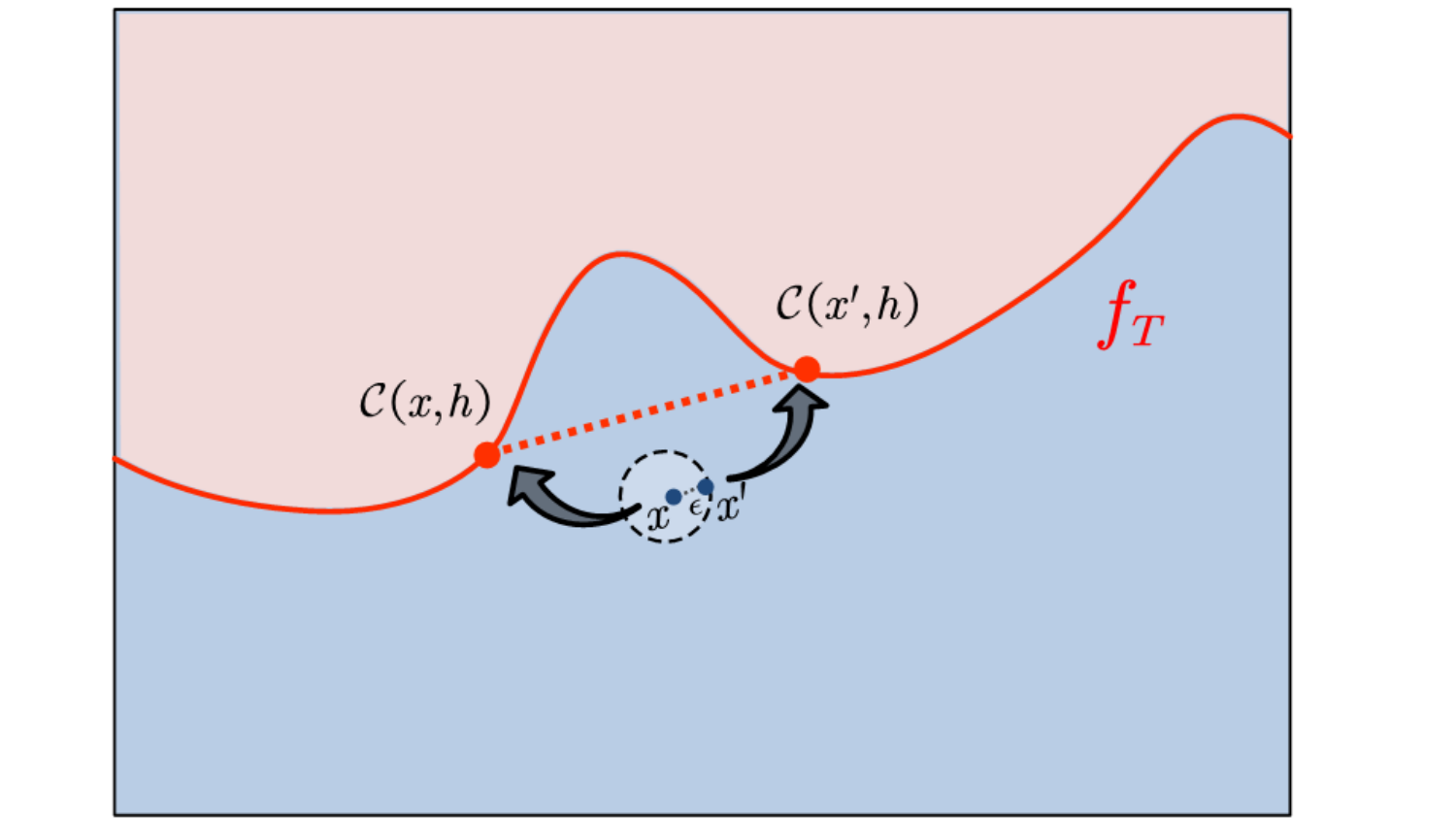}
    \caption{The shape of the decision boundary results in a substantial distance between generated counterfactuals regardless of how small the perturbation in $x$ may be. However, the discrepancy in explanatory strength remains relatively insignificant. Therefore, if robustness is assessed using Definition \ref{defSRC}, the accuracy and robustness of CEGAs are contradictory. Nevertheless, adopting Definition \ref{defWRC} for assessing robustness can circumvent such contradictions.}
    \label{fig:WRC}
\end{figure}
 Definition \ref{defSRC} closely aligns with our conventional understanding of robustness. Assuming that the current employed CEGA $\mathcal{C}$ is robust, when an input instance $x$ undergoes perturbation, the generation of the optimal counterfactual example corresponding to the output should exhibit only minor variations, as depicted in Figure \ref{fig:SRC}. 
\textbf{However, for CEGAs, such a definition can sometimes be biased.} As pointed out by Thibault Laugel et al.\cite{Laugel2019}, the classification boundary of classifiers trained using LAs may itself lead to non-robustness under the context of Definition \ref{defSRC}. In accordance with Figure \ref{fig:WRC}, the shape of theclassification boundary, denoted as $f_T$, leads to a scenario where even minute perturbations applied to the instance $x$ result in relatively significant alterations in the optimal counterfactual examples. Consequently, in such circumstances, the corresponding $\text{SRC}$ may exhibit substantial values. However, attributing this lack of robustness solely to CEGA $\mathcal{C}$ appears tenuous.
In a certain sense, Definition \ref{defSRC} imposes a somewhat stringent requirement on CEGAs. Therefore, we introduce the following definition:

\begin{definition}[Weak Robust Compatibility] \label{defWRC}
    Consider an instance space $(\mathcal{X},d)$, where  $\mathcal{X}$ is a subset of $K$-dimensional Euclidean space  $\mathbb{R}^K$, a LA represented as $L$, a CEGA as $\mathcal{C}$, $\varphi\in \Phi$.
    Let $D_T$ be a dataset of size $T$ drawn from the distribution $\mathcal{D}^{\otimes T}$, and $L(D_T)=h_T$.  We define the Weak Robust Compatibility (WRC) of  $\mathcal{C}$ w.r.t $h_T$ at $x$ as follows:
    \begin{align*}
    \textit{WRC}_x^{\varphi} (\mathcal{C}, h_T) = \int_{
     \mathcal{X}\cap B (x, r)} \tilde{\Delta}(x,y) \cdot \varphi (d (x, y)) d y ,
    \end{align*}
 where $B(x, r)$ denotes a hypersphere centered at $x $ with a radius of $r$ and 
     \begin{align*}
         \tilde{\Delta}(x,y)=\left | d (x, \mathcal{C} (h_T, x)) - d (y, \mathcal{C} (h_T, y))  \right | .
     \end{align*}
\end{definition}
\textbf{Comparison of Two Definitions.}
Compared to Definition \ref{defSRC}, Definition \ref{defWRC} no longer exclusively focuses on the robustness of the precise localization of the optimal counterfactual example for a given instance $x$. Reflecting upon the perspectives emphasized in Section \ref{sec:PreCE}, we can interpret 
$$d(x,\mathcal{C}(x,h_T)),d(y,\mathcal{C}(y,h_T))$$
as the counterfactual explanatory strength of CEGA $\mathcal{C}$ w.r.t $x,y$. Consequently, we can understand Definition \ref{defWRC} as the robustness of the counterfactual \textbf{explanatory strength}. It is evident that Definition \ref{defWRC} is applicable not only to scenarios illustrated in Figure \ref{fig:SRC} but also to situations resembling Figure \ref{fig:WRC}.
 In the subsequent sections of this paper, we shall exclusively focus on the concept of WRC.
 \begin{algorithm}[tb]
    \caption{Generating Robust Counterfactuals using WRC}
    \label{alg}
    \textbf{Input}: Model $h_T(\cdot)$, CEGA $\mathcal{C}(\cdot)$, instance $x$.\\
    \textbf{Parameter}: $\varphi(\cdot),r,\text{max\_steps},\tau>0,\sigma>0$.\\
    \textbf{Output}: The counterfactual of $x$.
    
    \begin{algorithmic}[1] 
        \STATE Let $\text{steps}=0$.
        \WHILE{$\overline{\text{WRC}}_x^{\varphi} (\mathcal{C}, h_T)\geq \tau$ and $\text{steps}< \text{max\_steps}$}
        \STATE Draw a random sample of $x'$ from the Gaussian distribution $\mathcal{N}(x,\sigma I_K)$. 
        \WHILE{$x'\notin B(x,r)$}
        \STATE Regenerate a sample x that follows the $\mathcal{N}(x,\sigma I_K)$.
        \ENDWHILE
        \STATE Let $x=x'$.
        \ENDWHILE
        \IF {$\text{steps}< \text{max\_steps}$}
        \STATE \textbf{return} Counterfactual $\mathcal{C}(x)$.
        \ELSE
        \STATE \textbf{return} No robust counterfactual found and exit. 
        \ENDIF
    \end{algorithmic}
\end{algorithm}
\subsection{Generating Robust Counterfactuals using WRC-Test}
Our proposed WRC is a generalization of robustness from the perspective of counterfactual explanatory strength. It can still help us generate more robust counterfactual explanations. We will present the specific experiments in Section \ref{sec:exp}. In this section, we introduce some necessary concepts required for the experiments. When considering the computation of WRC, we cannot directly use Definition \ref{defWRC}, but need to consider its discrete version: 
\begin{definition}[Discrete WRC]
With the notations defined in Definition \ref{defWRC}, we define the discrete WRC of $\mathcal{C}$ w.r.t $h_T$ at $x$ as follows:
    \begin{align*}
    \overline{\text{WRC}}_x^{\varphi} (\mathcal{C}, h_T) = \sum_{
    y\in\mathcal{N}_k(x) } 
    \tilde{\Delta}(x,y) \cdot \varphi (d (x, y)),
  \end{align*}
where $N_k(x),r$ is a set of $k$ points drawn form the uniform distribution $\mathcal{U}(B(x, r))$. 
\end{definition}
Next, we can define a roubustness test:
\begin{definition}[WRC-Test]\label{def:WRCtest}
    An instance $x\in\mathcal{X}$ satisfies the WRC-Test if:
    $$
    \overline{\text{WRC}}_x^{\varphi} (\mathcal{C}, h_T)\leq \tau.
  $$
\end{definition}
Naturally, we can use Definition \ref{def:WRCtest} to test the ability of an instance to generate robust counterfactual explanations. Therefore, when the current instance $x$ cannot generate sufficiently robust counterfactual explanations, we can consider searching for a substitute instance that is sufficiently close to $x$ and can pass the WRC-Test to generate counterfactual explanations. Specifically, inspired by \cite{pmlr-v202-hamman23a}, we propose the Algorithm \ref{alg}. It is noteworthy that in Algorithm \ref{alg}, we use Gaussian sampling to generate substitute samples that are close enough to the original instance x, making the substitution more reasonable.
\section{A Sufficient Condition for PAC WRC- Approximability}
\label{M}
\subsection{PAC WRC-Approximability}\label{subsec:PAC}

\begin{figure}
    \centering
    \includegraphics[width=1\linewidth]{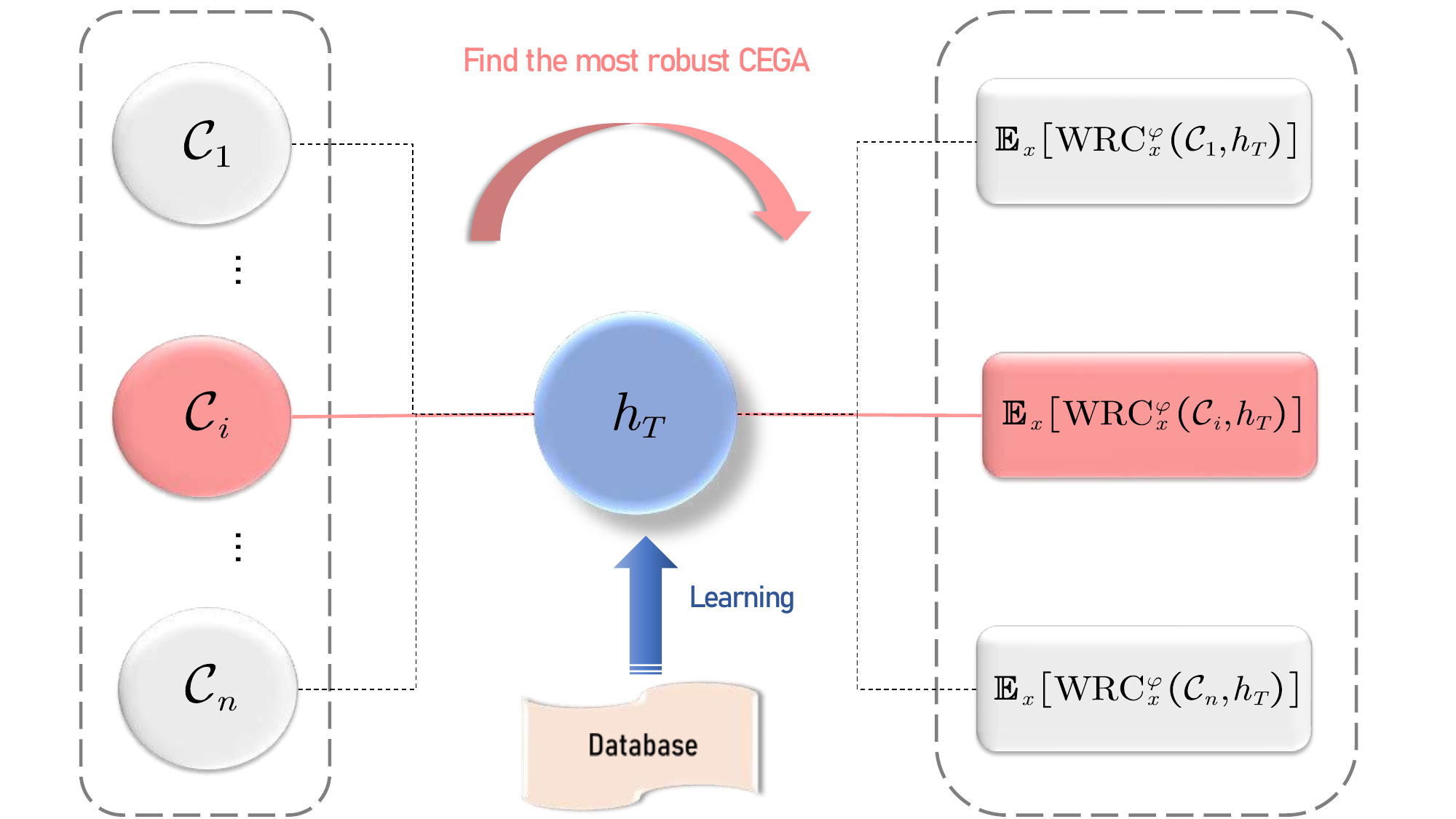}
    \caption{Another Application of WRC:  The numerically integrated WRC, estimated through computational methods, serves as a crucial reference metric for selecting a more robust CEGA.}
    \label{fig:WRCAPP}
\end{figure}

\textbf{Motivation: Another Application of WRC.} When we are not focused on a specific instance, WRC can also help us to choose a CEGA that is more robust to the current model. Given a model denoted as $h_T$ trained by $D_T,L$ along with several CEGAs denoted as $\mathcal{C}_1,\mathcal{C}_2,\cdots,\mathcal{C}_n$, if we seek to select a relatively robust-performing CEGA from this set, a reference approach is to employ numerical integration techniques such as Monte Carlo method \cite{Burden2015} to estimate $\mathbb{E}_x \left [ \textit{WRC}_x^{\varphi} (\mathcal{C}_1, h_T) \right ] ,
    \cdots,
    \mathbb{E}_x \left [ \textit{WRC}_x^{\varphi} (\mathcal{C}_n, h_T) \right ]$
    and subsequently choose the CEGA with the smallest value of this metric, as illustrated in Figure \ref{fig:WRCAPP}. However, in scenarios akin to online learning, where the dataset $D_T$ continues to expand and $h_T$ undergoes corresponding changes, calculating the WRC for a fixed $T$ is no longer as meaningful. In general, it is noteworthy that, as the volume of data continues to increase, the parameters of a trained model gradually converge. In other words, the classifier converges, i.e., $ h_T \rightarrow h^*$. At this point, $\mathbb{E}_x \left [ \textit{WRC}_x^{\varphi} (\mathcal{C}, h^*)\right ]$ remains a crucial reference, as it to some extent estimates the future trends in the WRC of  $\mathcal{C}$ and $L$. Admittedly, $h^*$ is unknown, making $\mathbb{E}_x \left [ \textit{WRC}_x^{\varphi} (\mathcal{C}, h^*)\right ]$ uncomputable directly. Consequently, we naturally contemplate the following question: \textbf{How large $T$ must be for} 
         $$\left | \mathbb{E}_x \left [ \textit{WRC}_x^{\varphi} (\mathcal{C}, h_T)\right ]-\mathbb{E}_x \left [ \textit{WRC}_x^{\varphi} (\mathcal{C}, h^*)\right ] \right | $$  
        \textbf{to be small enough?} 

\textbf{A Perspective from PAC Theory.}
In the field of statistical learning theory, a similar question arises: when $T$ is sufficiently large, how small is the difference between $R_{\mathcal{D}}(h_T)$ and $R_{\mathcal{D}}(h^*)$? A promising approach to addressing this question is to introduce the PAC framework \cite{mohri2018foundations,haussler1993probably}, which can be viewed as a rigorous mathematical analysis framework for machine learning. 
Within the PAC framework, if there exists a polynomial function $\text{poly}(1/\varepsilon,1/\delta,c)$ (where $c$ represents other relevant parameters of $\mathcal{H}$) such that for any $T>\text{poly}(1/\varepsilon,1/\delta,c)$, the following oracle inequality holds:
    \begin{align*}
        \Pr \left(R _{\mathcal{D}}(h_T) - R_{\mathcal{D}} (h^{\ast}) \leq \varepsilon\right) \geq 1 - \delta .
    \end{align*}
In this regard, we shall designate $\mathcal{H}$ as PAC learnable. The PAC learning theory has exerted a profound influence on the field of machine learning, contributing not only to theoretical research and algorithm design but also furnishing a profound understanding of machine learning problems, thereby offering crucial guidance for addressing practical issues. 

Now, We extend this classical mathematical framework to introduce the concept of PAC WRC-Approximable:
\begin{definition}[PAC WRC-Approximable] \label{defPAC}
    Let $T$ denote the number of samples independently and identically sampled from the distribution $\mathcal{D}$, where $0 < \varepsilon, \delta < 1$. For all distributions $\mathcal{D}$, if there exists a LA $L$, a CEGA $\mathcal{C}$, and a polynomial function $poly(1/\varepsilon, 1/\delta)$ such that the following inequality holds for any $ T\geq poly(1/\varepsilon, 1/\delta)$:
    \begin{align*}
    \Pr \left( \left | \mathbb{E}_x \left [ \textit{WRC}_x^{\varphi} (\mathcal{C}, h_T)\right ]-\mathbb{E}_x \left [ \textit{WRC}_x^{\varphi} (\mathcal{C}, h^*)\right ] \right |  \leq \varepsilon\right) \geq 1 - \delta .
    \end{align*}
    In this case, we say that $\mathcal{H}$ is PAC WRC-Approximable.
\end{definition}

 The remaining objective of Section \ref{M} is to provide a sufficient condition for $\mathcal{H}$ to be PAC WRC-Approximable. The organization of the remaining content is as follows: In Section \ref{subsec:Ass}, we will enumerate two plausible technical conditions. In Section \ref{subsec:M}, we will present our main results.
\subsection{Assumptions}\label{subsec:Ass}
In this section, we propose two reasonable assumptions, one for noise and the other for the classification boundary.

\textbf{Low Noise Condition.}
Tsybakov’s noise condition \cite{tsybakov2004}  delineates a crucial equilibrium between sample noise and model complexity. In numerous statistical estimation problems, the goal is to infer an unknown parameter from observed data, which is often subject to random noise. Tsybakov introduced this condition as a means to confine the intensity of noise, ensuring effective parameter or function estimation. Tsybakov’s noise condition has found widespread application across various domains, as evidenced by representative works \cite{wang2016noise,hanneke2011rates}.  Specifically, it assumes that there exists constants $c>0,0<\alpha\leq \infty$ such that
\begin{align}
\text{Pr}_{X\sim \mathcal{D}_{\mathcal{X} } }\left ( \big | \eta(X)-1/2\leq t  \big |  \right )  
\leq ct^{-\alpha },\label{eqTSY}
\end{align}
holds for all $0<t\leq t_0$, where $t_0$ is some constant and $$\eta(x)=\text{Pr}(Y=1|X=x).$$ 

\textbf{Smoothness.}
Assuming $f: \Omega \rightarrow \mathbb{R}$, where $\Omega\subset \mathbb{R}^d,\gamma > 0$. Let $\lfloor\gamma\rfloor$ be the largest integer less than $\gamma$. For any $d$-dimensional vector $\textbf{k} = (k_1,\cdots, k_d)$, we denote $|\textbf{k}| $as the sum of its components, i.e., $|\textbf{k}|=\sum_{i=1}^d k_i$, and 
$$
\partial^{\textbf{k}}=\partial^{|\textbf{k}|}/\partial^{k_1}x_1\cdots \partial^{k_d}x_d.
$$
Then, we define the set of $\gamma$-th order smooth functions as
$
C^{\gamma} (\Omega,R) := \{ f:\Omega\rightarrow \mathbb R : \| f \|_{\gamma} \leq R \},
$ 
where $\|f\|_\gamma$ \cite{wellner2013} represents
$$ \max_{| \textbf{k} | \leq \lfloor \gamma \rfloor} \|
\partial^{\textbf{k}} f (x) \|_{\infty} + \max_{| \textbf{k}| = \lfloor \gamma \rfloor} \sup_{x,
y \in \Omega} \frac{| \partial^{\textbf{k}} f (x) - \partial^{\textbf{k}} f (y) |}{\| x - y
\|^{\gamma - \lfloor \gamma \rfloor}} .$$
Given an integer $l \geq 1$, we consider a hypothesis set $\mathcal{H}_\gamma$ defined on the interval $[0,1]^{l+1}$. It satisfies that for all $h\in \mathcal{H}_\gamma$, the classification boundary of $h$ is defined as a $\gamma$-th order smooth function on \([0,1]^l\).
In a specific context, let $x=\textbf{x}_{1:l+1} = (x_1, \cdots,x_{l+1}) \in [0,1]^{l+1}$, where for all $h\in \mathcal{H}_\gamma$, it corresponds to a classification boundary, namely, the graph of the function $x_{l+1}=f(\textbf{x}_{1:l})$. We denote the Bayesian classifier in set $\mathcal{H}_\gamma$ as $h^*$, and the corresponding Bayesian classification boundary as $f^*$.
The VC dimension of the function class $C^{\gamma} (\Omega,R)$ is infinite, rendering it a function class of substantial statistical significance with favorable properties. Consequently, it finds extensive applications in various domains, such as asymptotic statistics and robust statistical learning \cite{xu2023non}. Meanwhile, in pursuit of obtaining a plethora of enlightening theoretical insights, we also contemplate the following classes of functions:
$$
C^{\infty} (\Omega,R) := \{ f:\Omega\rightarrow \mathbb R : \| f \|_{\gamma} \leq R ,\forall \gamma >0\}.
$$
The corresponding hypothesis space is denoted as $\mathcal{H}_\infty$ in our study.
\subsection{Theoretical Results}\label{subsec:M}
In this section, we present the main theoretical results of this paper. Due to space limitations, our proofs are provided in our supplementary material. Readers can refer to our supplementary materials for technical details. 

To begin, we state our first main theorem:
\begin{theorem}\label{thm1}
    Let $\mathcal H=\mathcal H_\gamma,\mathcal{X}=[0,1]^{l+1}$. Assume that the learning problem satisfies the Tsybakov’s noise condition (\ref{eqTSY}) and the CEGA $\mathcal{C}$ is induced by the metric $d(\cdot,\cdot)$ on $\mathcal{X}=[0,1]^{l+1}$. If hypothesis $h_T$ satisfies $R_{\mathcal{D}}(h_T)-R_{\mathcal{D}}\left(h^{*}\right)<\varepsilon$ holds with probability of at least $1-\delta/4$, then
    \begin{align*}
 \big | \mathbb{E}_x [ \textit{WRC}_x^{\varphi} (\mathcal{C}, h_T) ]-
\mathbb{E}_x[& 
 \textit{WRC}_x^{\varphi} (\mathcal{C}, h^*)]  \big| 
\\
&\leq \mathcal{O}\left(\varepsilon^{{\kappa\gamma} /(\gamma+l)}\right )
\end{align*}
    holds with probability of at least $1-\delta$.
\end{theorem}

Theorem \ref{thm1} reveals that, under certain conditions, the PAC bound can induce a bound for $\big | \mathbb{E}_x [ \textit{WRC}_x^{\varphi} (\mathcal{C}, h_T) ]-
\mathbb{E}_x[
 \textit{WRC}_x^{\varphi} (\mathcal{C}, h^*)]  \big| $. When considering $\mathcal H_\infty$, we can further establish the following theorem:
\begin{theorem}\label{thm2}
    Let $\mathcal H=\mathcal H_\infty,\mathcal{X}=[0,1]^{l+1}$. Assume that the learning problem satisfies the Tsybakov’s noise condition (\ref{eqTSY}) and the CEGA $\mathcal{C}$ is induced by the metric $d(\cdot,\cdot)$ on $\mathcal{X}=[0,1]^{l+1}$. If hypothesis $h_T$ satisfies $R_{\mathcal{D}}(h_T)-R_{\mathcal{D}}\left(h^{*}\right)<\varepsilon$ holds with probability of at least $1-\delta/4$, then
    \begin{align*}
 \big | \mathbb{E}_x [ \textit{WRC}_x^{\varphi} (\mathcal{C}, h_T) ]-
\mathbb{E}_x[ &
 \textit{WRC}_x^{\varphi} (\mathcal{C}, h^*)]  \big| 
\\
&\leq \mathcal{O}\left(\varepsilon^{\kappa }\cdot 
\left (  \log 1/\varepsilon  \right ) ^{2l} \right )
\end{align*}
    holds with probability of at least $1-\delta$.
\end{theorem}
Certainly, it is evident that Theorem \ref{thm1} (Theorem \ref{thm2}) provides a sufficient condition for $ \mathcal{H}_\gamma$ ($ \mathcal{H}_\infty$) to be PAC WRC-Approximable:
\begin{theorem}\label{cor1}
    Under the conditions of Theorem \ref{thm1}, a sufficient condition for $\mathcal{H}_\gamma$($ \mathcal{H}_\infty $) to be PAC WRC-Approximable is that $\mathcal{H}_\gamma$($ \mathcal{H}_\infty $) is PAC Learnable.
\end{theorem}
\textbf{Inspiration.} Theorem \ref{cor1} provides us with a profound insight, namely that under suitable conditions, the boundary of excess risk can determine the attainability of the WRC. In such circumstances, we can transform the non-intuitive problem of PAC-WRC Approximability into an estimation problem of excess risk. It is worth noting that in the field of statistical learning, extensive research has already been conducted on excess risk under various scenarios \cite{bartlett2006convexity,bartlett2002rademacher}.

\begin{table*}
    \renewcommand{\arraystretch}{1.1}
        \caption{We tested the metrics on DiCE and Proto-CF on different datasets using 100 randomly selected samples from the test set, the means of which are in the table, and the experimental results with the standard deviation will be shown in the Appendix.}
    \centering
    \scalebox{1}{  
    \begin{tabular}{clccccccccc}
    \hline
    & & \multicolumn{4}{c}{$l_1$ baesd} & &\multicolumn{4}{c}{$l_2$ based}\\
    \cline{3-6} \cline{8-11}
       & Method  
       & \textbf{COST} & \textbf{LOF} & \textbf{WRC} & \textbf{VAL} &
       & \textbf{COST} & \textbf{LOF} & \textbf{WRC} & \textbf{VAL} \\
    \hline
     \multirow{4}{*}{ \rotatebox{90}{HELOC}}    
        & DiCE 
        & 0.38 & 0.82 & 2.73 & $85\%$ & 
        & 0.36 & 0.80 & 3.34 & $92\%$ \\   
        & DiCE+WRC-Test (Ours)
        & 0.48 & 0.88 & 1.37 & $91\%$ & 
        & 0.42 & 0.84 & 1.72 & $98\%$ \\
        & Proto-CF
        & 0.25 & 1.00 & 0.05 & $100\%$ & 
        & 0.31 & 1.00 & 0.07 & $100\%$ \\
        & Proto-CF +WRC-Test (Ours)
        & 0.27 & 1.00 & 0.04 & $100\%$  & 
        & 0.30 & 1.00 & 0.04 & $100\%$ \\

    \hline
         \multirow{4}{*}{\rotatebox{90}{GERMAN}}  
        & DiCE 
        & 1.28 & 0.88 & 1.35 & $97\%$ & 
        & 1.17 & 0.90 & 1.49 & $93\%$ \\
        & DiCE+WRC-Test (Ours)
        & 1.37 & 0.92 & 0.92 & $94\%$ & 
        & 1.22 & 0.96 & 1.01 & $95\%$ \\
        & Proto-CF
        & 0.46 & 0.96 & 0.08 & $100\%$ & 
        & 0.59 & 0.94 & 0.12 & $100\%$ \\
        & Proto-CF +WRC-Test (Ours)
        & 0.45 & 0.98 & 0.04 & $100\%$  & 
        & 0.59 & 0.98 & 0.07 & $100\%$ \\
        
    \hline
         \multirow{4}{*}{\rotatebox{90}{ADULT}}  
         & DiCE 
        & 0.32 & 0.86 & 2.71 & $87\%$ & 
        & 0.28 & 0.90 & 3.12 & $89\%$ \\
        & DiCE+WRC-Test (Ours)
        & 0.37 & 0.90 & 1.67 & $92\%$ & 
        & 0.29 & 0.96 & 1.82 & $95\%$ \\
        & Proto-CF
        & 0.06 & 1.00 & 0.08 & $100\%$ & 
        & 0.06 & 1.00 & 0.11 & $100\%$ \\
        & Proto-CF +WRC-Test (Ours)
        & 0.07 & 1.00 & 0.05 & $100\%$  & 
        & 0.06 & 1.00 & 0.06 & $100\%$ \\

    \hline
         \multirow{4}{*}{\rotatebox{90}{COMPAS}}  
        & DiCE 
        & 1.48 & 0.84 & 3.29 & $90\%$ & 
        & 1.59 & 0.82 & 2.75 & $91\%$ \\
        & DiCE+WRC-Test (Ours)
        & 1.42 & 0.90 & 1.17 & $92\%$ & 
        & 1.61 & 0.90 & 1.09 & $95\%$ \\
        & Proto-CF
        & 0.03 & 0.98 & 0.07 & $100\%$ & 
        & 0.05 & 1.00 & 0.12 & $100\%$ \\
        & Proto-CF +WRC-Test (Ours)
        & 0.05 & 1.00 & 0.02 & $100\%$  & 
        & 0.06 & 1.00 & 0.05 & $100\%$ \\
       
    \hline
    \end{tabular}
    }
    \label{exp1table}
\end{table*}

\section{Experiments}\label{sec:exp}
In this section, we present experimental results to demonstrate the effectiveness of Algorithm \ref{alg}. 
\subsection{Experimental Setup}
In this section, we outline our experimental setup.

\textbf{Datasets.}
Our experiments are implemented on four binary classification datasets: 
\begin{itemize}
    \item HELOC \cite{Heloc}: From the NeurIPS 2017 Explainable Machine Learning Challenge, this dataset predicts loan defaults using customer financial data.
    \item Adult \cite{misc_adult_2}: Aimed at predicting whether individuals earn over $\$50,000$ per year, utilizing demographic data.
    \item German Credit \cite{misc_statlog_(german_credit_data)_144}: Assesses credit risk by classifying individuals into good or bad credit risks, based on their financial and personal information.
    \item COMPAS \cite{kaggle_compas}: Focuses on predicting the likelihood of reoffending after prison release, using demographic and criminal history features.
\end{itemize}

\textbf{Performance Metrics.}
Our metrics of interest are: 
\begin{itemize}
    \item Cost: Average $l_1$ or $l_2$ distance between counterfactuals $x'$ and original points $x$ (lower is better).
    \item Validity($\%$): The CEGA gives the percentage of counterfactual instances that have the same label as the input instance (higher is better).
    \item LOF: A measure of whether the counterfactual explanation is on the data manifold (higher is better).
    \item WRC: The metrics we defined in Section \ref{subsecWRC} to measure the stability of counterfactual explanations (lower is better). 
\end{itemize}

\textbf{CEGAs.} 
We utilized the WRC method to find more robust counterfactual explanations based on two open-source counterfactual explanation generation algorithms.
\begin{itemize}
    \item DiCE \cite{mothilal2020dice}: 
DiCE utilizes a distance loss for continuous features, which is weighted by the absolute median deviation, and an indicator function loss for discrete features. This approach compares input and output instances. By optimizing the sum of these two losses, it achieves the optimal counterfactual explanation.
    \item Proto-CF \cite{van2021interpretable}: Proto-CF add another term to the loss that optimizes for the distance between the counterfactual instance and representative members of the target class. In doing this, Proto-CF require interpretability also to mean that the generated counterfactual is believably a member of the target class and not just in the data distribution.
\end{itemize}

\textbf{Methodology.}
Before starting the experiments, we normalized the datasets to weaken the importance difference of different features. Then we designed a three-layer fully connected neural network model with ReLU activation function for each processed dataset and ensured that the accuracy of the trained network model is above 70$\%$, and our subsequent comparison experiments will be conducted on the processed dataset and the trained model to ensure fairness.

\textbf{Hyperparameter Selection.}
Due to the different number of features in different datasets, the performance of the two counterfactual explanation methods on the neural network model is different, the parameter $\tau$ is set to the values $(1.8, 1.1, 2.0, 1.3)$ for DiCE,  $(0.05, 0.05, 0.07, 0.06)$ for Proto-CF, on the four respective datasets. The maximum number of search steps is denoted as $\text{max}\_\text{steps}=30$, and the sampling radius is defined as $r=10^{-5}\cdot n$, where $n$ represents the number of features.
We employed the existing algorithms in Scikit-learn to compute the LOF metric, with the parameter "kneighbors" configured to a value of $10$. Simultaneously, we opt for the function $\varphi(x)=1/(x+\varepsilon)$ where $\varepsilon=10^{-6}$.

\subsection{Experiment Results}
Table \ref{exp1table} presents the performance of two different CEGAs, DiCE and Proto-CF, on four datasets and neural network models, both before and after integration with the WRC-Test. From the perspective of VAL, DiCE's performance is inferior to Proto-CF on all four datasets. Notably, DiCE exhibits lower VAL values when dealing with neural network models of poorer performance. In contrast, Proto-CF consistently achieves an impressive 100$\%$ VAL in the experiments. Regarding the COST metric, Proto-CF has a lower COST compared to DiCE. In terms of robustness indicators LOF and WRC, Proto-CF also outperforms DiCE. The robustness indicators LOF and WRC improve significantly for both DiCE and Proto-CF after the integration with WRC-Test. Naturally, the process of generating replacement samples in Algorithm \ref{alg} might lead to an increase in the COST metric. However, in certain scenarios with high robustness requirements, a moderate sacrifice in COST is acceptable.
In summary, counterfactuals generated by the existing CEGAs exhibit favorable cost characteristics; however, the robustness of the counterfactuals they produce may be insufficient. The integration of our proposed WRC-Test, indeed, proves beneficial in assisting users in identifying more robust counterfactual explanations.
\section{Conclusion}
\label{Con}
In this paper, addressing the concerns raised by Thibault Laugel et al. \cite{Laugel2019}, we introduce a more principled concept of WRC: it no longer relies solely on the perspective of perturbation instances but, rather, relaxes the constraints on the shape of the classification boundary from the standpoint of explanatory strength. Consequently, it encompasses the characteristics of both CEGAs and LAs, rendering it more widely applicable. 
In practice, we proposed the WRC-Test and designed experiments to verify its rationality. 
Theoretically, for a reasonable and secure estimation of $\mathbb{E}_x[
 \textit{WRC}_x^{\varphi} (\mathcal{C}, h^*)]$, we incorporate insights from PAC learning theory and define the concept of PAC WRC-Approximability. Under reasonable assumptions, we establish oracle inequalities concerning WRC, which provides a sufficient condition for PAC WRC-Approximability, thus contributing to its theoretical significance.

\textbf{Open Problems.} In the theoretical section of our work, one of our key assumptions is the smoothness assumption on the hypothesis space $\mathcal{H}$. An intriguing open question arises: if the hypotheses in hypothesis space $\mathcal{H}$ no longer adhere to the smoothness assumption for classification boundaries but merely satisfy the continuity assumption, how should we investigate PAC WRC-Approximability? A more challenging problem emerges when the hypotheses in hypothesis space $\mathcal{H}$ no longer yield classification boundaries expressible by functions but are defined by algebraic equations representing curves. How should we then delve into the study of PAC WRC-Approximability? These are all meaningful and formidable questions.
\section*{Acknowledgement}
This work is supported by the National Key Research and Development Program of China (Grant No.2022YFB3103702).
\appendix

\bibliographystyle{named}
\bibliography{ijcai24}

\end{document}